\documentclass[]{style/ceurart}

\usepackage{listings}
\usepackage{array}
\begin{document}

\copyrightyear{2024}
\copyrightclause{Copyright for this paper by its authors.
  Use permitted under Creative Commons License Attribution 4.0
  International (CC BY 4.0).}

\conference{CLEF 2024: Conference and Labs of the Evaluation Forum, September 09-12, 2024, Grenoble, France}

\title{Using Pretrained Large Language Model with Prompt Engineering to Answer Biomedical Questions}
\subtitle{DS@GT CLEF2024 BioASQ Task 12b and Synergy Task Working Note}

\author[1]{Wenxin Zhou}[
orcid=0009-0002-3325-3357,
email=wzhou77@gatech.edu,
]
\author[2]{Thuy Hang Ngo}[
email=tngo62@gatech.edu,
]
\cormark[1]

\address[1]{Georgia Institute of Technology, North Ave NW, Atlanta, GA 30332, United States}
\cortext[1]{Corresponding author.}

\begin{abstract}
Our team participated in the BioASQ 2024 Task12b and Synergy tasks to build a system that can answer biomedical questions by retrieving relevant articles and snippets from the PubMed database and generating exact and ideal answers.
We propose a two-level information retrieval and question-answering system based on pre-trained large language models (LLM), focused on LLM prompt engineering and response post-processing.
We construct prompts with in-context few-shot examples and utilize post-processing techniques like resampling and malformed response detection.
We compare the performance of various pre-trained LLM models on this challenge, including Mixtral, OpenAI GPT and Llama2.
Our best-performing system achieved 0.14 MAP score on document retrieval, 0.05 MAP score on snippet retrieval, 0.96 F1 score for yes/no questions, 0.38 MRR score for factoid questions and 0.50 F1 score for list questions in Task 12b.
\end{abstract}

\begin{keywords}
  large language model \sep
  prompt engineering \sep
  biomedical information retrieval \sep
  biomedical question answering
\end{keywords}

\maketitle

\section{Introduction}

BioASQ is a challenge for large-scale biomedical semantic indexing and question answering hosted by CLEF.
The BioASQ12b and the Synergy tasks \cite{task12bSynergy2024overview} are part of the CLEF 2024 BioASQ lab\cite{BioASQ2024overview}, which focuses on biomedical question answering and information retrieval.
The challenge consists of four types of questions: yes/no, factoid, list and summary. The participating systems need to perform two subtasks.

The first subtask is to retrieve 10 relevant documents and snippets from the PubMed database that can answer the question. PubMed \cite{pubmedoverview} is a search engine for biomedical literature, which contains millions of abstracts of biomedical articles. The system is evaluated by the relevance of the retrieved documents and snippets using the mean average precision (MAP) metric.

The second subtask is to generate an exact answer and an ideal answer for each question. The exact answer is a short answer that directly answers the question. For yes/no questions, this is a single word "yes" or "no". For list and factoid questions, the short answer is a list of entities. The ideal answer is a long answer that provides more context and details. The system is evaluated based on the quality and accuracy of the generated answers. The evaluation metric is F1 score for yes/no questions, mean reciprocal rank (MRR) for factoid questions and F1 score for list questions. The ideal answer is scored manually based on the readability, recall, precision and repetition of the answers.
\begin{figure}
    \centering
    \includegraphics[width=\linewidth]{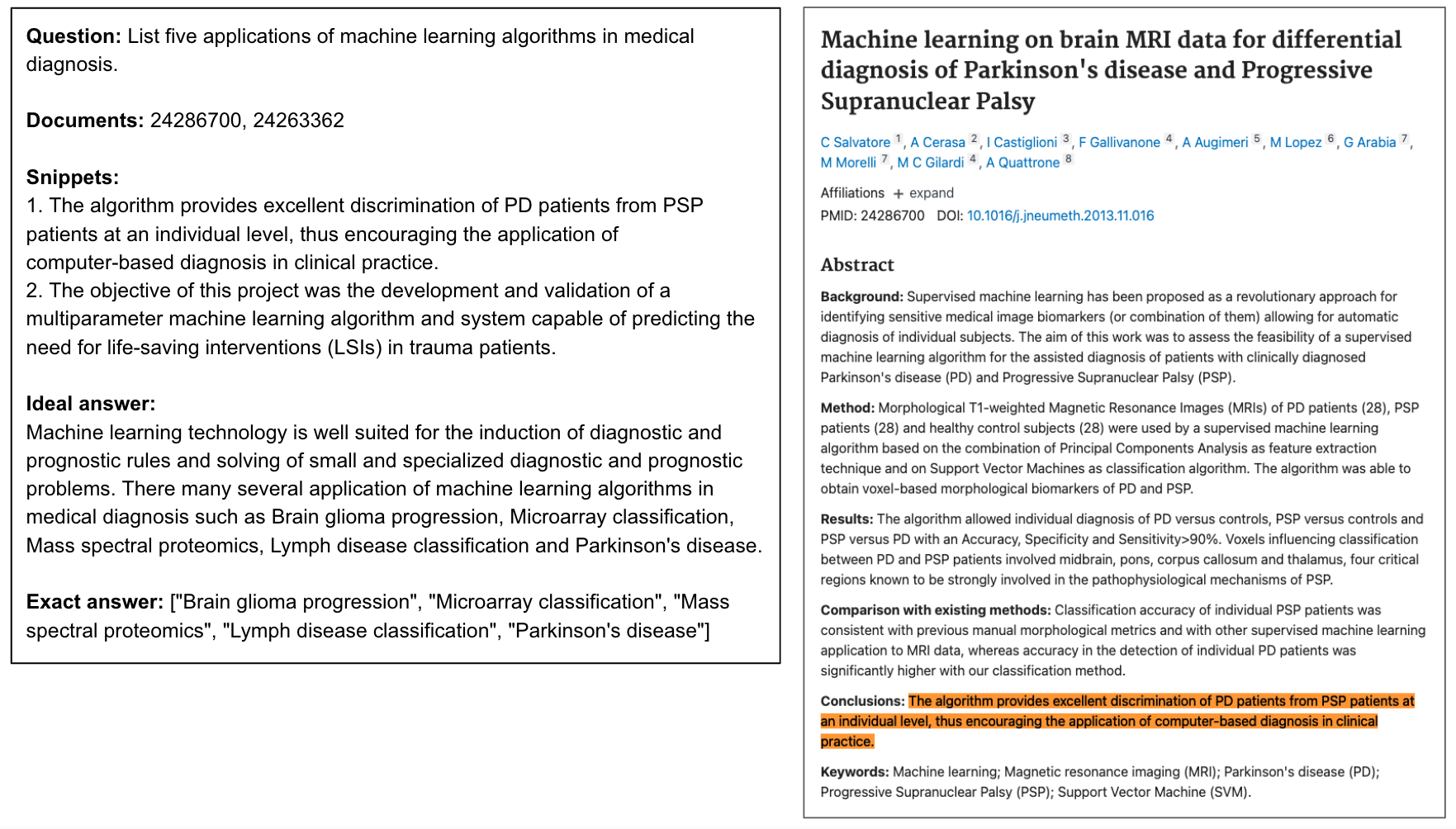}
    \caption{BioASQ Data Format}
    \label{fig:data_format}
\end{figure}

An example of the input and output format is shown in Figure~\ref{fig:data_format}. The organizers provide BioASQ-QA dataset\cite{BioASQQADataSet}, which contains around 4721 questions from the past BioASQ challenges where 27\% are yes/no questions, 29\% factoid, 24\% summary and 20\% list.

We build a system based on pre-trained large language models for document retrieval and question-answering. Although some solutions of previous years used large language models, they only experimented with OpenAI GPT models and basic prompt engineering. In this year's challenge, we experiment with various well-known large language models and use prompt engineering and response post-processing techniques to improve the performance of the system. At high level, we use LLM model to extract keywords from the question and compose PubMed query to retrieve documents from PubMed database, then use sentence embeddings to find the relevant snippets from the documents. For question answering, we use the snippets as context and construct few-shot examples prompts to guide the LLM to generate the answers in the desired format.
In this paper, we will discuss the modeling pipeline, prompt engineering strategies as well as the experiment results with various LLM models on the Synergy and Task12b tasks. Our implementation can be found on Github\footnote{{https://github.com/dsgt-kaggle-clef/bioasq-2024/}}.


\section{Related Work}

Large language models (LLM) have shown great success recently in various natural language processing tasks, including text generation in the biomedical domain.
\citet{chatgptBLURB} measured the performance of LLM on Biomedical Language Understanding and Reasoning Benchmark (BLURB), demonstrating the potential of LLM in understanding and reasoning in the biomedical domain.
Prompt engineering \cite{promptengineering} is a technique that improves the performance of LLM for domain specific tasks.
In-context few-shot examples in the prompt can help LLM to generate more accurate answers, without the need for fine-tuning the model.
Some well-known LLM models include OpenAI GPT \cite{chatgpt4}, Meta Llama2 \cite{llama2}, and Mistral AI's Mixtral \cite{mixtral} models.

\subsection{Information Retrieval Approaches in BioASQ 2023}

In the eleventh BioASQ challenge, two predominant methodologies were employed for the Information Retrieval (IR) task, typically segmented into a two-stage pipeline: retrieval and reranking.

For the retrieval stage, the majority of the systems (7 out of 8) involved in Task 11B phase A adopted a BM25 model for the initial document retrieval \cite{BioASQ2023Overview}. BM25 models rely on indexing the entire corpus of documents which is computationally expensive since more than a million biomedical papers are added to the PubMed database each year \cite{informationoverload}. The advantage of this method is the comprehensive list of documents that could be retrieved. 

In contrast, \citet{isgptexpert} utilized LLMs for retrieval via zero-shot learning by mirroring the expert's workflow in curating the BioASQ QA dataset. This method does not require additional computation to index the document corpus, but may not be as comprehensive as the BM25 method. The system did not achieve the highest IR outcomes in last year's competition.

Most participating systems generated language model embeddings for reranking articles. This process was handled by a dedicated reranking model or through cosine similarity measures \cite{BioASQ2023Overview}. One system employed zero-shot prompting with an LLM to rank the documents, but was resource-intensive and limited by the maximum context length of the model.

\subsection{Question Answering Approaches in BioASQ 2023}

For the Question Answering (QA) or Phase B task, several systems exploited the capabilities of LLMs by prompting it with essential snippets of information. A comparative analysis revealed that GPT-4 outperformed an ensemble of fine-tuned BERT models, indicating that GPT-4 is more effective at navigating the complexities of biomedical question answering \cite{BiomedicalQAApproaches}.

\section{Methodology}

\subsection{Information Retrieval}
\begin{figure}
    \centering
    \includegraphics[width=0.9\linewidth]{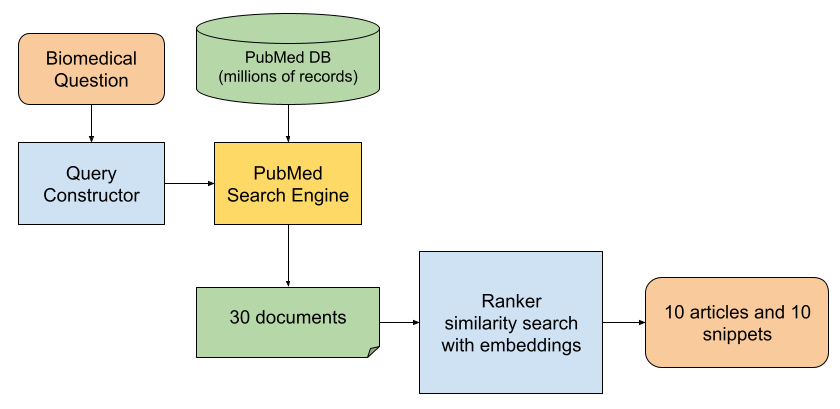}
    \caption{Information Retrieval System Design}
    \label{fig:ir_design}
\end{figure}

We propose a two-stage IR system for the BioASQ task (in Figure~\ref{fig:ir_design}). The first stage retrieves a set of relevant documents from the PubMed database using PubMed search query \cite{pubmedquery}. The second stage ranks documents by cosine similarity of sentence embeddings to find the most relevant sentences.

\subsubsection{Query Constructor}

The query constructor creates the query for PubMed search \cite{pubmedoverview}. To match the PubMed version defined by the organizer, we set the {\it maxdate} parameter in the {\it esearch} API to be the date defined by the specific batch of the Synergy task. Task12b requires PubMed 2024 baseline, so we set the {\it maxdate} parameter to 2024-01-01 as an approximation. The query constructor uses two approaches.

\textbf{Approach 1: Keyword extraction.} We use LLMs or language models finetuned for biomedical terminology to extract keywords (such as biomedical entities) from the question. Then we concatenate those keywords with "AND" to form a PubMed query. For LLMs, we send a few-shot example prompt, shown in Table~\ref{Tab:irprompt1} to generate the keywords. For the biomedical language model, we use {\it en\_ner\_bc5cdr\_md}, a spaCy biomedical NER named entity recognition language model trained on BC5CDR corpus \cite{scispacy} to extract the keywords from the question sentence.
\begin{table}[ht]
    \caption{Query Constructor Prompt, Approach 1}
    \centering
    \fbox{\begin{minipage}{0.98\linewidth}
        Q: What is the mode of action of Molnupiravir? \\
        Keywords: Molnupiravir, action \\
        \#\#\#\\
        Q: Is dapagliflozin effective for COVID-19?\\
        Keywords: dapagliflozin, COVID-19\\
        \#\#\#\\
        Q: \{question body\}\\
        Keywords:
    \end{minipage}}
    \label{Tab:irprompt1}
\end{table} 

\textbf{Approach 2: Direct query generation.} We directly generate a query from the question using the large language model, which is inspired by Ateia and  Kruschwitz \cite{isgptexpert}. The prompt template is composed of an instruction and two examples, as shown in Table \ref{Tab:irprompt2}.
\begin{table}[ht]
    \caption{Query Constructor Prompt, Approach 2}
    \centering
    \fbox{\begin{minipage}{0.98\linewidth}
        Given a question, expand into a search query for PubMed by incorporating synonyms and additional terms that would yield relevant search results from PubMed to the provided question while not being too restrictive.
        Assume that phrases are not stemmed; therefore, generate useful variations. Return only the query that can directly be used without any explanation text.\\

        Question: What is the mode of action of Molnupiravir? \\
        Query: Molnupiravir AND ("mode of action" OR mechanism) \\
        \#\#\#\\
        Question: Is dapagliflozin effective for COVID-19? \\
        Query: dapagliflozin AND (COVID-19 OR SARS-CoV-2 OR coronavirus) AND (efficacy OR effective OR treatment) \\
        \#\#\#\\
        Question: Name monoclonal antibody against SLAMF7. \\
        Query: "SLAMF7" AND ("monoclonal antibody" OR "monoclonal antibodies") \\
        \#\#\#\\
        Question: \{body\} \\
        Query:
    \end{minipage}}
    \label{Tab:irprompt2}
\end{table}

\subsubsection{Reranker}

The reranker ranks documents by calculating the relevance between documents and questions. We use a sentence transformer, specifically {\it all-MiniLM-L6-v2} \cite{minilm} to generate embeddings for the documents and the questions. When the document length is larger than the maximum input length of the sentence transformer, the document is truncated to fit the input length. We calculate the cosine similarity between the embeddings of the question and the document and use the descending score to rank the documents.

\subsubsection{Snippet Extraction}

After identifying the top 10 documents, we break the documents into sentences and rank the sentences based on the similarity score using
the same sentence transformer and similarity calculation method as the re-ranker. We then select the best sentence of each document as the prediction
snippets.

\subsection{Question Answering}

\begin{figure}
    \centering
    \includegraphics[width=\linewidth]{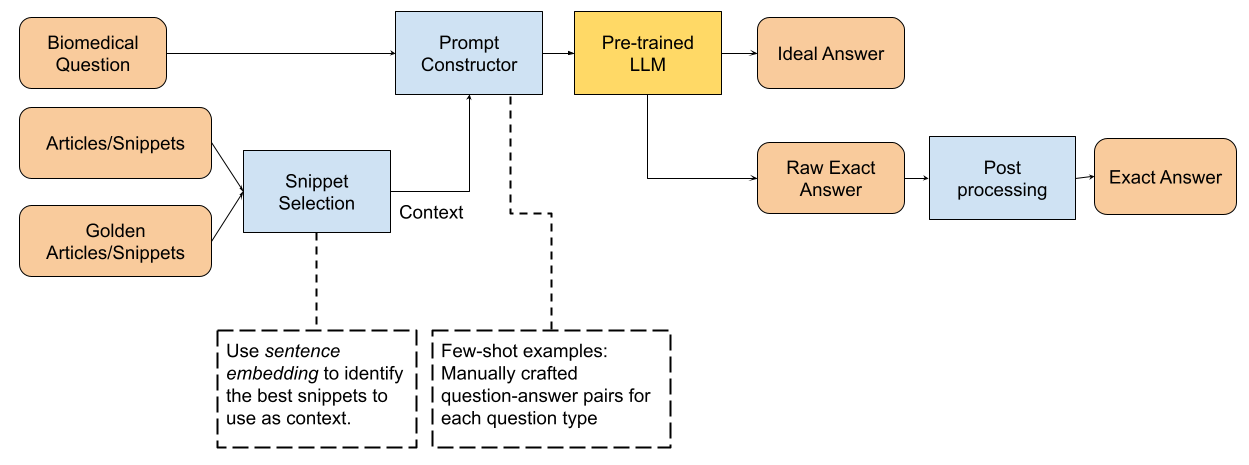}
    \caption{Question Answering System Design}
    \label{fig:qa_design}
\end{figure}

We use pre-trained large language models (LLM) to generate answers for biomedical questions. In this project, instead of fine-tuning the LLM models, we use prompt engineering and response post-processing to build the system, as illustrated in Figure~\ref{fig:qa_design}. The two key components of the prompt are the context and the few-shot examples.

We use the first 1000 words of the top 10 snippets for the question, where each snippet consists of one or more sentences of a PubMed abstract, as the context for the question answering. The snippets are generated from the information retrieval system or are the golden snippets provided by the organizers.
The reasoning behind using the first 1000 words of snippets is that the higher rank snippets contain the more relevant information to the question. The context is crucial for generating high-quality answers and reducing model hallucinations.
Then we construct few-shot examples from the training dataset. The few-shot examples help LLM to generate answers in the desired format. The template prompt used for yes/no questions is shown in Table \ref{Tab:qatemplate}. The maximum input token size of most LLM models we experimented with is larger than 4096, so a prompt consisting of few-shot examples, 1000 words (which is roughly equivalent to 1350 tokens) context as well as the question body is within the LLM input token limit.

\begin{table}[ht]
    \caption{Prompt Template for Question Answering, Yes/no Question}
    \centering
    \fbox{\begin{minipage}{0.98\linewidth}
        \textbf{Context}: Papilins are homologous, secreted extracellular matrix proteins which share a common order of protein domains. \\
        \textbf{Question}: Is the protein Papilin secreted? \\
        \textbf{Ideal answer}: Yes, papilin is a secreted protein \\
        \textbf{Exact answer}: yes \\
        \#\#\#\\
        \textbf{Context}: Most lncRNAs are under lower sequence constraints than protein-coding genes and lack conserved secondary structures, making it hard to predict them computationally.\\
        \textbf{Question}: Are long non coding RNAs as conserved in sequence as protein coding genes?\\
        \textbf{Ideal answer}: No. Most long non coding RNAs are under lower sequence constraints than protein-coding genes.\\
        \textbf{Exact answer}: no\\
        \#\#\#\\
        \textbf{Context}: \{context\}\\
        \textbf{Question}: \{question\_body\}\\
        \textbf{Ideal answer}:
    \end{minipage}}
    \label{Tab:qatemplate}
\end{table}

The prompt templates used for factoid, list and summary questions are similar to the yes/no question template with some modifications to the ideal and exact answer formatting. Those templates can be found in the appendix \ref{appendix:prompt_template}.

"\#\#\#" is used as the separator for examples in the prompt, which is also used as the "stop" string of the LLM completion.
The prompt is intentionally designed to end with "Ideal answer:" to guide the LLM to generate the ideal answer. We expect the "exact answer" line to be generated after the ideal answer in the LLM response
as illustrated in appendix \ref{appendix:prompt_response}.
In terms of the other parameters (such as temperature and top\_k) for the LLM completion, we use the default values defined by the TextSynth service \cite{textsynth}, except that the {\it max\_tokens} parameter, which controls maximum number of tokens in the LLM output is set to 200. We did not experiment with using different roles or system and user prompts since not all models we experimented with are fined-tuned for chat, so we only relied on the basic completion functionality of LLM to generate the answers.

For list questions, we experiment with synonym grouping. The idea is to group the synonyms among all the LLM responses to reduce the repetition of the answers. This is similar to having LLM perform a second-stage reasoning. The synonym grouping is accomplished by sending a prompt (shown in Table \ref{Tab:groupingprompt}) to LLM to group the synonyms. The prompt contains a list of entities, which aggregates all entities returned from multiple responses from LLM for the list question with different prompt contexts.

\begin{table}[ht]
    \caption{Prompt Template for Synonym Grouping}
    \centering
    \fbox{\begin{minipage}{0.98\linewidth}
        Group the phrases with the same meaning in the ENTITY list into separate lines as follows.\\
        \lbrack ENTITY\rbrack: MOG-IgG; AQP4; MOG-IgG; serum neurofilament light chain; NfL; aquaporin-4 (AQP4)-immunoglobulin G (IgG)\\
        \lbrack GROUP1\rbrack: aquaporin-4 (AQP4)-immunoglobulin G (IgG); AQP4; MOG-IgG \\
        \lbrack GROUP2\rbrack: serum neurofilament light chain; NfL\\
        \#\#\#\\
        \lbrack ENTITY\rbrack: \{entity\_list\}\\
        \lbrack GROUP1\rbrack: \\
    \end{minipage}}
    \label{Tab:groupingprompt}
\end{table}

\subsubsection{Context formation}

We started with the first 1000 words of the top 10 snippets as the prompt context. The system we submit to Synergy uses this basic version. We experiment with different contexts for the QA system by changing the number and variety of snippets used. Our final context setup used for batch 2 and 3 in Task 12b is as follows:

\begin{enumerate}
    \item For yes/no questions, we create three prompts with different contexts.
The context of the first prompt is the first golden snippet, the context of the second prompt is the second golden snippet, and the context of the third prompt is the third golden snippet.
We send all three prompts separately to the LLM and take the majority vote of the answers as the final answer.
    \item For factoid and summary questions, we use one prompt with the first 1000 words of the golden snippets as the context.
    \item For list questions, we use one prompt with the first 1000 words of the golden snippets as the context.
    In the synonym grouping setting, we compose five prompts. The context of each prompt is a golden snippet. The first prompt context is the first golden snippet, the second prompt context is the second golden snippet, and so on.
\end{enumerate}

\subsubsection{Response Post-processing}

Since the prompt we send to LLM has examples with the desired answer format, the answers generated by LLM are usually in the form of a long answer,
followed by an exact answer in the second line. We extract answers by parsing the response.
For list and factoid questions we give examples where entities are separated by semicolons, therefore we extract resulting entities by splitting the exact answer string by semicolons.
Examples of prompt and response can be found in appendix \ref{appendix:prompt_response}.
When the answer does not follow the expected format, we resample the LLM output. Some checks we use to detect malformed answers detection are:

\begin{itemize}
    \item There is no "exact answer" string in the response for yes/no, factoid and list questions.
    \item The exact answer is not "yes" or "no" for yes/no questions.
    \item The exact answer for list or factor questions separates entities by commas or newlines instead of semicolons.
\end{itemize}

\section{Results}

In this section, we present and discuss the results of our systems for the BioASQ Task 12b and Synergy tasks. The results are based on the official evaluation scores provided by the organizers in the competition leaderboard for BioASQ2024.

\subsection{Synergy Task}

We submitted five systems for the Synergy task to measure the performance of different pre-trained large language models and strategies. The system configurations are outlined in Table~\ref{Tab:synergy_task_config}.

\begin{table}[ht]
    \caption{Synergy Task System Configuration}
    \begin{tabular}{| m{1.5cm} | m{2.5cm}| m{3.5cm} |m{2cm}| m{1.5cm} |}
        \hline
        System & Leaderboard Name & IR algorithm & QA Model & QA has context\\
        \hline\hline
        system1 & Gatech competition & spaCy biomedical NER model & Mistral 7B & No \\
        \hline
        system2 & GTBioASQsys2 & LLM (Mistral 7B) + few-shot prompt & Mistral 7B & Yes \\
        \hline
        system3 & GTBioASQsys3 & LLM (Mixtral 47B) + few-shot prompt & Mixtral 47B Instruct & Yes \\
        \hline
        system4 & GTBioASQsys4 & LLM (Llama2 70B) + few-shot prompt & Llama2 70B & Yes \\
        \hline
        system5 & GTBioASQsys5 & LLM (GPT-J 6B) + few-shot prompt & GPT-J 6B & Yes \\
        \hline
    \end{tabular}
    \label{Tab:synergy_task_config}
\end{table}

For the information retrieval (IR) part, system 1 uses {\it en\_ner\_bc5cdr\_md} to extract the keywords from the questions. The rest of the systems use large language generative models (LLM) to extract the keywords.
The language models used for systems 2,3,4,5 are Mistral 7B, Mixtral 47B (i.e, Mixtral 8x7B model) \cite{mixtral}, llama2 \cite{llama2}, GPT-J \cite{gpt-j} respectively.

For the question-answering part, the prompt of system 1 contains no context, whereas the rest of the systems use snippets as the context.
The LLM models used for systems 1,2,3,4,5 are Mistral 7B, Mistral 7B, Mixtral 47B, llama2 and GPT-J respectively.

\begin{table}[ht]
    \caption{Synergy Round 4 Information Retrieval Results}
    \begin{tabular}{| m{2.5cm} | m{2cm}| m{2cm} |m{2cm}| m{2cm} |}
        \hline
        System & Document Macro-F1 & Document MAP & Snippet Macro-F1 & Snippet MAP \\
        \hline\hline
        top competitor & {\bf 0.1459} & {\bf 0.1308} & {\bf 0.0982} & {\bf 0.1003} \\
        \hline
        system1 & 0.001 & 0.0003 & 0.0014 & 0.0003 \\
        \hline
        system2 & 0.0283 & 0.027 & 0.0146 & 0.0215 \\
        \hline
        system3 & {\bf0.0578} & {\bf0.0434} & {\bf0.034} & {\bf0.0317} \\
        \hline
        system4 & 0.0355 & 0.0258 & 0.0225 & 0.0249 \\
        \hline
        system5 & 0.0353 & 0.0267 & 0.0273 & 0.0237 \\
        \hline
    \end{tabular}
    \label{Tab:synergy_ir_results}
\end{table}

Table~\ref{Tab:synergy_ir_results} shows the information retrieval results for Synergy task round 4.
System 3 has the best performance with 0.0434 mean-average precision (MAP) score for document retrieval and 0.031 for snippet retrieval. The performance of systems 2,4 and 5 is similar in MAP score, in the range between 0.02 and 0.03, whereas system 1 is the worst with MAP score of 0.0003. The systems used for the Synergy task only perform the basic first-level retrieval by fetching 10 records from PubMed using a query concatenated by keywords.
We can see that Mixtral47B outperforms other systems for the question keyword extraction task.
The spaCy language model {\it en\_ner\_bc5cdr\_md} performs the worst. The reason is that the {\it en\_ner\_bc5cdr\_md} model is often unable to detect any keywords in the question body since it is limited to detecting only the disease and chemical entities in the sentence.

\begin{table}[ht]
    \caption{Synergy Round 4 Question Answering Results}
    \begin{tabular}{| m{2.5cm} | m{2cm}| m{2cm} |m{2cm}| m{3cm} |}
        \hline
        System & Yes/no F1 & Factoid MRR & List F1 & Ideal Answer Manual Score (total out of 20) \\
        \hline\hline
        top competitor & {\bf 1.0} & {\bf 0.43} & {\bf 0.27} & {\bf 17.41} \\
        \hline
        system1 & 0.33 & 0.1818 & 0.0806 & 13.56  \\
        \hline
        system2 & 0.4 & {\bf0.3182} & 0.2009 & 14.91 \\
        \hline
        system3 & {\bf0.625} & 0.2727 & {\bf0.2256} & {\bf16.4} \\
        \hline
        system4 & 0.4857 & 0.2727 & 0.1919 & 11.55 \\
        \hline
        system5 & 0.4857 & 0.1364 & 0.1467 & 9.05 \\
        \hline
    \end{tabular}
    \label{Tab:synergy_qa_results}
\end{table}

Table~\ref{Tab:synergy_qa_results} shows the question-answering results for Synergy task round 4. Our best-performing system achieved 0.62 F1 score for yes/no questions, 0.3182 on mean reciprocal rank (MRR) metric for factoid questions and 0.22 F1 score for list questions.
The Mixtral47B model gives the best performance on all question types except for factoid questions. For factoid questions, system 2 with Mistral 7B model performs the best.
Even though the Llama2 model is much larger in model size, it does not perform as well as the Mixtral/Mistral models.
By comparing the results of system1 and system2, we can see that adding context to the prompt greatly improves the QA answering accuracy for all types of questions for the same LLM model engine.

The ideal answer is evaluated by manual scores according to the readability, recall, precision and repetition of the answers.
Mixtral 47B performs the best in terms of generating long answers. The Llama2 model performs the worst in generating long answers, even though it is on par with other models in exact answer generation.
It seems to indicate that the Llama2 model is not good at generating long answers, which could be due to the model architecture differences.
It is also possible that the Llama2 model was trained to give vague answers to biomedical questions such that it does not mislead the users on health issues.

\subsection{BioASQ Task 12B}
For Task 12B, we added the direct query generation method to our experiment.
We updated the re-ranker component to get the top 10 documents among the top 30 documents retrieved from the first stage via the PubMed Query for system 1 in batch3.
In addition, we enhanced the system by response resampling and adding a fallback to use a query with the original question if LLM keyword extraction fails to generate any keyword or the query generated by GPT-4 \cite{chatgpt4} returns no results.

\subsubsection{Task 12B Phase A}
For the three systems we submit to the PhaseA of task 12B, system 1 uses the direct query generation method with GPT-3.5 for batch 1 and GPT-4 for batch 2 and 3. Systems 2 and 3 continue to use the keyword extraction method with Mistral 7B and Mixtral 47B as before. The system configurations are outlined in Table~\ref{Tab:12b_system}.
\begin{table}[ht]
    \caption{12B System Configurations}
    \begin{tabular}{| m{1.3cm} | m{2.3cm}| m{3.5cm} |m{2cm}| m{3.0cm} |}
        \hline
        System & Leader-board Name & Phase A IR method & Phase A+ QA context & Phase B QA strategy\\
        \hline\hline
        system1 & Gatech competition & Query generation with GPT3.5 or GPT-4 & IR snippet & Golden snippet (add synonym grouping for list question in batch 2 and 3) \\
        \hline
        system2 & GTBioASQsys2 & Keyword extraction with Mistral 7B & No snippet & Golden snippet \\
        \hline
        system3 & GTBioASQsys3 & Keyword extraction with Mixtral 47B & IR snippet & - \\
        \hline
    \end{tabular}
    \label{Tab:12b_system}
\end{table}

In Table~\ref{Tab:12b_ir_results}, all three systems in batch 2 have similar performance in terms of MAP score at around 0.081 for document retrieval, with system 3 having the best performance.
For snippet retrieval, system 3 has the best performance with MAP score of 0.0271, followed by system 2 and system 1. The system 1 performance also improved from 0.0497 in batch 1 to 0.081 in batch 2, after switching from using GPT-3.5 to GPT-4 for query generation.

In batch 3, system 1 had a significant improvement of MAP to 0.1385 thanks to increasing the number of articles retrieved from PubMed in the initial retrieval stage from 10 to 30. This allows for more articles to be processed in the reranking stage and results in higher recall overall. In past competitions, solutions that use BM25 models for retrieval fetch hundreds of documents in the initial stage of retrieval \cite{domainspecificbert}, these systems also tend to have the best score for IR task. We hypothesize that our systems, which use LLM for the retrieval stage, would have even better performance should the number of articles retrieved initially be increased further to 100. However, due to the time required to fetch the articles from PubMed API, extract the snippets, and score the articles for similarity with the query, increasing the number of retrieved document results in a long wait time.

\begin{table}[ht]
    \caption{12B Phase A Information Retrieval Results}
    \begin{tabular}{| m{1.5cm} | m{3.5cm} | m{2.5cm}| m{2.5cm} |}
        \hline
        Batch & System & Document MAP & Snippet MAP \\
        \hline\hline
        \multirow{4}{*}{1} & top competitor & {\bf 0.2067} & {\bf 0.1149} \\
        & system1 & 0.0497 & 0.0150 \\
        & system2 & 0.0599 & 0.0050\\
        & system3 & {\bf0.0717} & {\bf0.0204} \\
        \hline
        \multirow{4}{*}{2} & top competitor & {\bf 0.2293} & {\bf 0.1539} \\
         & system1 & 0.081 & 0.0151 \\
         & system2 & 0.0826 & 0.0261\\
         & system3 & {\bf0.0836} & {\bf0.0271} \\
        \hline
        \multirow{4}{*}{3} & top competitor & {\bf 0.2549}  & {\bf 0.2224} \\
        & system1 & {\bf0.1385} & {\bf0.0505} \\
        & system2 & 0.0883 & 0.0312\\
        & system3 & 0.0929 & 0.0271 \\
        \hline
        \end{tabular}
    \label{Tab:12b_ir_results}
\end{table}

\subsubsection{Task 12B Phase A+ and Phase B}

For the question answering (QA) part, we enhanced our system by adding resampling if the exact answer does not satisfy the requirements. For example, neither "yes" nor "no" is in the answer for yes/no question.
We also experimented with setting up different contexts for the QA system, by changing the number of snippets and the variety of the snippets used.

We submit three systems to PhaseA+. Phase A+ system1 uses the snippets generated by system1 in PhaseA as the context for the QA prompt. Phase A+ system 2 does not use any snippet as the context for the QA prompt. Phase A+ system 3 uses the snippets generated by system 3 in PhaseA as the context.
In PhaseB, we use the golden snippet provided by the organizer as the context of the QA prompt for PhaseB system1 and system2. The difference is that PhaseB system1 performs synonym grouping for list questions, whereas PhaseB system2 does not.

Table~\ref{Tab:12b_qa_results} shows the results of all the five systems in PhaseA+ and PhaseB.
Take batch 2 as an example, the system without context (Phase A+ system2) only achieved 0.69 F1 score for yes/no questions.
Adding context improves the F1 score to 0.80 for yes/no questions and adding golden snippets as the context further improves the F1 score to 0.96 for yes/no questions.
For factoid questions, adding non-golden snippets as context does not improve the MRR score, but adding golden snippets as context improves the MRR score from 0.21 to 0.36.
For list questions, adding non-golden snippets as context improves the F1 scores slightly, and adding golden snippets as context further improves the F1 score from 0.21 to 0.50.
Batch 1 and 3 results also follow the same pattern.

\begin{table}[ht]
    \caption{12B Question Answering Results}
    \begin{tabular}{| m{2cm} | m{3.5cm} | m{2cm}| m{2cm} |m{2cm}|}
        \hline
        Batch & System & Yes/no F1 & Factoid MRR & List F1 \\
        \hline\hline
        \multirow{6}{*}{1} & Phase A+ top competitor & 0.9167 & 0.3611 & 0.4808\\
        &Phase A+ system1 & 0.8397 & 0.1429 & 0.3661  \\
        &Phase A+ system2 & 0.7987 & 0.0952 & 0.2350 \\
        &Phase A+ system3 & 0.8397 & 0.1429 & 0.1866 \\
        &Phase B top competitor & {\bf0.9589} & {\bf0.4286} & {\bf0.5843} \\
        &Phase B system1 & {\bf0.8792} & {\bf0.1905} & {\bf0.3975} \\
        \hline
        \multirow{7}{*}{2} & Phase A+ top competitor & 0.9601 & 0.3947 & 0.4088 \\
        & Phase A+ system1 & 0.8051 & 0.2105 & 0.2133  \\
        & Phase A+ system2 & 0.6905 & 0.2105 & 0.1268 \\
        & Phase A+ system3 & 0.8003 & 0.2105 & 0.1364 \\
        & Phase B top competitor & {\bf0.9601} & {\bf0.6842} & {\bf0.5708} \\
        & Phase B system1 & {\bf0.9601} & {\bf0.3684} & 0.2936 \\
        & Phase B system2 & 0.8802 & 0.358 & {\bf0.4964} \\
        \hline
        \multirow{7}{*}{3} & Phase A+ top competitor & 0.9143 & 0.3301 & 0.3708 \\
        & Phase A+ system1 & 0.7822 & 0.2308 & 0.1774  \\
        & Phase A+ system2 & 0.6643 & 0.0769 & 0.2098 \\
        & Phase A+ system3 & 0.7483 & 0.1538 & 0.1860 \\
        & Phase B top competitor & {\bf 1.0} & {\bf 0.5} & {\bf 0.6484} \\
        & Phase B system1 & 0.8748 & 0.3077 & 0.3399 \\
        & Phase B system2 & {\bf0.9577} & {\bf0.3846} & {\bf0.4763} \\
        \hline
        \end{tabular}
    \label{Tab:12b_qa_results}
\end{table}

\section{Discussion}
The Task12B results show that our systems with golden snippets as the context can achieve an F1 score of 0.87-0.96 for yes/no questions.
We improved post-processing steps for factoid and list questions after batch1, by removing duplicate answers and detecting malformed answers.
Therefore the MRR score of our final system for factoid question is in the range of 0.3-0.4. The F1 score for list question is 0.45-0.5.

By comparing the list F1 score of PhaseB system1 and system2 in batch 2 and 3, we can see that synonym grouping performs worse than not using synonym grouping.
To understand the reasons, we looked at some of the synonym grouping responses from LLM and found that LLM often groups entities that should be in different categories together. Table \ref{Tab:groupingsample} shows an example prompt and response pair. We can see that  "fibromyalgia" and "chronic fatigue syndrome" are grouped as synonyms, and "depression" and "hypermobility spectrum disorders" are grouped as synonyms, whereas they should all be separate entities.
As a result, the synonym grouping does not help the list question performance. This also demonstrated that adding second-stage reasoning using LLM does not always give better results for complex problems.

\begin{table}[ht]
    \caption{Example Response from LLM to Synonym Grouping Prompt}
    \centering
    \begin{flushleft}
        \textbf{Grouping Prompt:}
    \end{flushleft}
    \fbox{\begin{minipage}{0.98\linewidth}
        Group the phrases with the same meaning in the ENTITY list into separate lines as follows.\\
        (examples)\\
        \lbrack ENTITY\rbrack: fibromyalgia; chronic fatigue syndrome; autosomal dominant polycystic kidney disease; Marfan syndrome; Cutis laxa syndromes; depression; osteogenesis Imperfecta Type 1; Loey-Dietz syndrome; hypermobility spectrum disorders\\
        \lbrack GROUP1\rbrack: \\
    \end{minipage}}

    \begin{flushleft}
        \textbf{Grouping Response:}
    \end{flushleft}
    \fbox{\begin{minipage}{0.98\linewidth}
        fibromyalgia; chronic fatigue syndrome \\
        \lbrack GROUP2\rbrack: autosomal dominant polycystic kidney disease; Marfan syndrome; osteogenesis Imperfecta Type 1; Loey-Dietz syndrome \\
        \lbrack GROUP3\rbrack: Cutis laxa syndromes \\
        \lbrack GROUP4\rbrack: depression; hypermobility spectrum disorders \\
    \end{minipage}}
    \label{Tab:groupingsample}
\end{table}

Our key takeaways from the experiments are:
\begin{enumerate}
    \item For IR part, Mixtral 47B is the best-performing model for question keyword extraction among the models we have tested. Retrieving more documents in the initial retrieval stage can improve the performance of the system.
    \item For QA part, adding context, especially using "correct" snippets as the context, to the prompt can greatly improve the QA answering accuracy for all types of questions.
    \item The improvements of the QA scores in batch 2 and 3 in Task 12B demonstrate that resampling LLM response is a great technique to improve accuracy. Simple response post-processing steps to validate the output format can also improve the performance.
    \item By comparing the results of Llama2 and other models in the Synergy task, we found that Llama2 model is not good at generating long answers, even though it is on par with other models in exact answer generation. 
    \item Two-stage LLM reasoning does not always give better results for complex problems, as shown by the synonym grouping experiment.
\end{enumerate}

\section{Future Work}
Here are some ideas for future work to improve the performance of our systems.

For the IR part, currently, we only fetch a small amount of documents from PubMed and use embeddings to rank the documents. We found that increasing the number of documents fetched in the initial retrieval stage improves the recall and overall MAP score but leads to long processing time. Calculating embeddings on the fly is especially time-consuming.
In the future, we want to embed all the PubMed documents in advance and store the embeddings in a vector database. In this way, we can fetch more documents in the first stage retrieval for second stage reranking as we would be able to look up the embeddings for a specific document quickly. We can also use similarity search on the vector database to directly fetch relevant documents for the question.

When calculating the similarity between the question and the document, we only use the first part of the document, which fits the embedding model input token size. We want to investigate if
splitting the documents into multiple parts and calculating the similarity for each part can improve the performance. We can also explore the performance of using different sentence embedding models.

For the QA part, our current system is based on the few-shot examples to guide the LLM to generate answers. We only used a few training examples in the BioASQ dataset and have not utilized the potential of
the BioASQ dataset. The next step would be to fine-tune the pretrained LLM model (specifically Mixtral47B) on the BioASQ dataset. The experience of
crafting examples for prompt engineering can help us prepare training data for fine-tuning LLM.
We will consider using Low Rank Adaptation (LoRA) \cite{hu2022lora} as a cost-effective method for finetuning a model with a large number of parameters.

\section{Conclusions}

We implemented information retrieval and question-answering systems for the BioASQ Task 12b and Synergy tasks.
The information retrieval system uses pretrained LLM and prompt engineering to search documents and uses sentence embeddings to rank documents.
The question answering system uses in-context few-shot examples to guide the LLM to generate answers while passing article snippets as context.
Our final system incorporates several useful techniques such as resampling and response post-processing for LLM interaction.
We experimented with various state-of-the-art LLM models, compared their performance and found that Mixtral 47B is overall the best-performing model.
Our best-performing system achieved 0.14 MAP score on document retrieval, 0.05 MAP score on snippet retrieval,
0.96 F1 score for yes/no questions, 0.38 MRR score for factoid questions, and 0.50 F1 score for list questions in Task 12b.
We hope this work can provide insights for future research in building biomedical question answering systems using large language models.

\section*{Acknowledgements}
Thank you to the Data Science @ Georgia Tech (DS@GT CLEF) team and Anthony Miyaguchi for their support.
We acknowledge the use of Grammarly \cite{grammarly} to proofread this paper.

\bibliography{main}

\begin{thebibliography}{21}
\expandafter\ifx\csname natexlab\endcsname\relax\def\natexlab#1{#1}\fi
\providecommand{\url}[1]{\texttt{#1}}
\providecommand{\href}[2]{#2}
\providecommand{\path}[1]{#1}
\providecommand{\DOIprefix}{doi:}
\providecommand{\ArXivprefix}{arXiv:}
\providecommand{\URLprefix}{URL: }
\providecommand{\Pubmedprefix}{pmid:}
\providecommand{\doi}[1]{\href{http://dx.doi.org/#1}{\path{#1}}}
\providecommand{\Pubmed}[1]{\href{pmid:#1}{\path{#1}}}
\providecommand{\bibinfo}[2]{#2}
\ifx\xfnm\relax \def\xfnm[#1]{\unskip,\space#1}\fi
\bibitem[{Nentidis et~al.(2024{\natexlab{a}})Nentidis, Katsimpras, Krithara, and Paliouras}]{task12bSynergy2024overview}
\bibinfo{author}{A.~Nentidis}, \bibinfo{author}{G.~Katsimpras}, \bibinfo{author}{A.~Krithara}, \bibinfo{author}{G.~Paliouras},
\newblock \bibinfo{title}{{Overview of BioASQ Tasks 12b and Synergy12 in CLEF2024}},
\newblock in: \bibinfo{editor}{G.~Faggioli}, \bibinfo{editor}{N.~Ferro}, \bibinfo{editor}{P.~Galuščáková}, \bibinfo{editor}{A.~García Seco~de Herrera} (Eds.), \bibinfo{booktitle}{Working Notes of CLEF 2024 - Conference and Labs of the Evaluation Forum}, \bibinfo{year}{2024}{\natexlab{a}}.
\bibitem[{Nentidis et~al.(2024{\natexlab{b}})Nentidis, Katsimpras, Krithara, Lima-López, Farré-Maduell, Krallinger, Loukachevitch, Davydova, Tutubalina, and Paliouras}]{BioASQ2024overview}
\bibinfo{author}{A.~Nentidis}, \bibinfo{author}{G.~Katsimpras}, \bibinfo{author}{A.~Krithara}, \bibinfo{author}{S.~Lima-López}, \bibinfo{author}{E.~Farré-Maduell}, \bibinfo{author}{M.~Krallinger}, \bibinfo{author}{N.~Loukachevitch}, \bibinfo{author}{V.~Davydova}, \bibinfo{author}{E.~Tutubalina}, \bibinfo{author}{G.~Paliouras},
\newblock \bibinfo{title}{{Overview of BioASQ 2024: The twelfth BioASQ challenge on Large-Scale Biomedical Semantic Indexing and Question Answering}},
\newblock in: \bibinfo{editor}{L.~Goeuriot}, \bibinfo{editor}{P.~Mulhem}, \bibinfo{editor}{G.~Quénot}, \bibinfo{editor}{D.~Schwab}, \bibinfo{editor}{L.~Soulier}, \bibinfo{editor}{G.~Maria Di~Nunzio}, \bibinfo{editor}{P.~Galuščáková}, \bibinfo{editor}{A.~García Seco~de Herrera}, \bibinfo{editor}{G.~Faggioli}, \bibinfo{editor}{N.~Ferro} (Eds.), \bibinfo{booktitle}{Experimental IR Meets Multilinguality, Multimodality, and Interaction. Proceedings of the Fifteenth International Conference of the CLEF Association (CLEF 2024)}, \bibinfo{year}{2024}{\natexlab{b}}.
\bibitem[{pub(2023)}]{pubmedoverview}
\bibinfo{title}{Pubmed overview}  (\bibinfo{year}{2023}). \URLprefix \url{https://pubmed.ncbi.nlm.nih.gov/about/}.
\bibitem[{Krithara et~al.(2023)Krithara, Nentidis, Bougiatiotis, and Paliouras}]{BioASQQADataSet}
\bibinfo{author}{A.~Krithara}, \bibinfo{author}{A.~Nentidis}, \bibinfo{author}{K.~Bougiatiotis}, \bibinfo{author}{G.~Paliouras},
\newblock \bibinfo{title}{{BioASQ-QA: A manually curated corpus for Biomedical Question Answering}},
\newblock \bibinfo{journal}{Scientific Data} \bibinfo{volume}{10} (\bibinfo{year}{2023}) \bibinfo{pages}{170}.
\bibitem[{Chen et~al.(2023)Chen, Sun, Liu, Jiang, Ran, Jin, Xiao, Lin, Chen, and Niu}]{chatgptBLURB}
\bibinfo{author}{Q.~Chen}, \bibinfo{author}{H.~Sun}, \bibinfo{author}{H.~Liu}, \bibinfo{author}{Y.~Jiang}, \bibinfo{author}{T.~Ran}, \bibinfo{author}{X.~Jin}, \bibinfo{author}{X.~Xiao}, \bibinfo{author}{Z.~Lin}, \bibinfo{author}{H.~Chen}, \bibinfo{author}{Z.~Niu},
\newblock \bibinfo{title}{{An extensive benchmark study on biomedical text generation and mining with ChatGPT}},
\newblock \bibinfo{journal}{Bioinformatics} \bibinfo{volume}{39} (\bibinfo{year}{2023}) \bibinfo{pages}{btad557}. \URLprefix \url{https://doi.org/10.1093/bioinformatics/btad557}. \DOIprefix\doi{10.1093/bioinformatics/btad557}.
\bibitem[{White et~al.(2023)White, Fu, Hays, Sandborn, Olea, Gilbert, Elnashar, Spencer-Smith, and Schmidt}]{promptengineering}
\bibinfo{author}{J.~White}, \bibinfo{author}{Q.~Fu}, \bibinfo{author}{S.~Hays}, \bibinfo{author}{M.~Sandborn}, \bibinfo{author}{C.~Olea}, \bibinfo{author}{H.~Gilbert}, \bibinfo{author}{A.~Elnashar}, \bibinfo{author}{J.~Spencer-Smith}, \bibinfo{author}{D.~Schmidt}, \bibinfo{title}{A prompt pattern catalog to enhance prompt engineering with chatgpt}, \bibinfo{year}{2023}. \DOIprefix\doi{10.48550/arXiv.2302.11382}.
\bibitem[{OpenAI(2023)}]{chatgpt4}
\bibinfo{author}{OpenAI},
\newblock \bibinfo{title}{Gpt-4 a large-scale transformer-based language model},
\newblock \bibinfo{journal}{OpenAI}  (\bibinfo{year}{2023}). \URLprefix \url{https://chat.openai.com}.
\bibitem[{Touvron et~al.(2023)Touvron, Lavril, Izacard, Martinet, Lachaux, Lacroix, Rozière, and Goyal}]{llama2}
\bibinfo{author}{H.~Touvron}, \bibinfo{author}{T.~Lavril}, \bibinfo{author}{G.~Izacard}, \bibinfo{author}{X.~Martinet}, \bibinfo{author}{M.-A. Lachaux}, \bibinfo{author}{T.~Lacroix}, \bibinfo{author}{B.~Rozière}, \bibinfo{author}{N.~Goyal},
\newblock \bibinfo{title}{Llama open and efficient foundation language models},
\newblock \bibinfo{journal}{arXiv preprint arXiv:2302.13971}  (\bibinfo{year}{2023}). \URLprefix \url{https://arxiv.org/abs/2302.13971}.
\bibitem[{Jiang et~al.(2024)Jiang, Sablayrolles, Roux, Mensch, Savary, Bamford, and Chaplot}]{mixtral}
\bibinfo{author}{A.~Q. Jiang}, \bibinfo{author}{A.~Sablayrolles}, \bibinfo{author}{A.~Roux}, \bibinfo{author}{A.~Mensch}, \bibinfo{author}{B.~Savary}, \bibinfo{author}{C.~Bamford}, \bibinfo{author}{D.~S. Chaplot},
\newblock \bibinfo{title}{Mixtral of experts},
\newblock \bibinfo{journal}{arXiv preprint arXiv:2401.04088}  (\bibinfo{year}{2024}). \URLprefix \url{https://arxiv.org/abs/2401.04088}.
\bibitem[{Nentidis et~al.(2023)Nentidis, Katsimpras, Krithara, López, Farré-Maduell, Gasco, Krallinger, and Paliouras}]{BioASQ2023Overview}
\bibinfo{author}{A.~Nentidis}, \bibinfo{author}{G.~Katsimpras}, \bibinfo{author}{A.~Krithara}, \bibinfo{author}{S.~L. López}, \bibinfo{author}{E.~Farré-Maduell}, \bibinfo{author}{L.~Gasco}, \bibinfo{author}{M.~Krallinger}, \bibinfo{author}{G.~Paliouras},
\newblock \bibinfo{title}{Overview of bioasq 2023: The eleventh bioasq challenge on large-scale biomedical semantic indexing and question answering}  (\bibinfo{year}{2023}). \URLprefix \url{https://arxiv.org/abs/2307.05131}.
\bibitem[{Landhuis(2016)}]{informationoverload}
\bibinfo{author}{E.~Landhuis},
\newblock \bibinfo{title}{Scientific literature: Information overload},
\newblock \bibinfo{journal}{Nature} \bibinfo{volume}{535} (\bibinfo{year}{2016}) \bibinfo{pages}{457--458}. \URLprefix \url{https://www.nature.com/articles/nj7612-457a}. \DOIprefix\doi{https://doi.org/10.1038/nj7612-457a}, \bibinfo{note}{published online 20 July 2016}.
\bibitem[{Ateia and Kruschwitz(2023)}]{isgptexpert}
\bibinfo{author}{S.~Ateia}, \bibinfo{author}{U.~Kruschwitz},
\newblock \bibinfo{title}{Is chatgpt a biomedical expert? exploring the zero-shot performance of current gpt models in biomedical tasks},
\newblock \bibinfo{journal}{CEUR Workshop Proceedings} \bibinfo{volume}{3497} (\bibinfo{year}{2023}). \URLprefix \url{https://ceur-ws.org/Vol-3497/paper-006.pdf}.
\bibitem[{Kim et~al.(2023)Kim, Hwang, Lee, Minju~Seo, and Kang}]{BiomedicalQAApproaches}
\bibinfo{author}{H.~Kim}, \bibinfo{author}{H.~Hwang}, \bibinfo{author}{C.~Lee}, \bibinfo{author}{W.~Y. Minju~Seo}, \bibinfo{author}{J.~Kang},
\newblock \bibinfo{title}{Exploring approaches to answer biomedical questions: From pre-processing to gpt-4}  (\bibinfo{year}{2023}). \URLprefix \url{https://ceur-ws.org/Vol-3497/paper-011.pdf}.
\bibitem[{Sayers(2009)}]{pubmedquery}
\bibinfo{author}{E.~Sayers},
\newblock \bibinfo{title}{A general introduction to the e-utilities},
\newblock \bibinfo{journal}{National Center for Biotechnology Information (US)}  (\bibinfo{year}{2009}). \URLprefix \url{https://www.ncbi.nlm.nih.gov/books/NBK25497/}.
\bibitem[{Neumann et~al.(2019)Neumann, King, Beltagy, and Ammar}]{scispacy}
\bibinfo{author}{M.~Neumann}, \bibinfo{author}{D.~King}, \bibinfo{author}{I.~Beltagy}, \bibinfo{author}{W.~Ammar},
\newblock \bibinfo{title}{{S}cispa{C}y: {F}ast and {R}obust {M}odels for {B}iomedical {N}atural {L}anguage {P}rocessing}  (\bibinfo{year}{2019}) \bibinfo{pages}{319--327}. \URLprefix \url{https://www.aclweb.org/anthology/W19-5034}. \DOIprefix\doi{10.18653/v1/W19-5034}. \href{http://arxiv.org/abs/arXiv:1902.07669}{{\tt arXiv:arXiv:1902.07669}}.
\bibitem[{Face(2021)}]{minilm}
\bibinfo{author}{H.~Face},
\newblock \bibinfo{title}{sentence-transformers/all-minilm-l6-v2},
\newblock \bibinfo{journal}{Hugging Face Community week}  (\bibinfo{year}{2021}). \URLprefix \url{https://huggingface.co/sentence-transformers/all-MiniLM-L6-v2}.
\bibitem[{tex(2024)}]{textsynth}
\bibinfo{title}{Textsynth documentation}  (\bibinfo{year}{2024}). \URLprefix \url{https://textsynth.com/documentation.html}.
\bibitem[{Wang and Komatsuzaki(2021)}]{gpt-j}
\bibinfo{author}{B.~Wang}, \bibinfo{author}{A.~Komatsuzaki},
\newblock \bibinfo{title}{{GPT-J-6B: A 6 Billion Parameter Autoregressive Language Model}}  (\bibinfo{year}{2021}).
\bibitem[{Maël~Lesavourey(2023)}]{domainspecificbert}
\bibinfo{author}{G.~H. Maël~Lesavourey},
\newblock \bibinfo{title}{Bioasq 11b: Integrating domain specific vocabulary bert-based model for biomedical document reranking},
\newblock \bibinfo{journal}{Working Notes of the Conference and Labs of the Evaluation Forum (CLEF 2023)}  (\bibinfo{year}{2023}). \URLprefix \url{https://ceur-ws.org/Vol-3497/}.
\bibitem[{Hu et~al.(2022)Hu, Shen, Wallis, Allen-Zhu, Li, Wang, Wang, and Chen}]{hu2022lora}
\bibinfo{author}{E.~J. Hu}, \bibinfo{author}{Y.~Shen}, \bibinfo{author}{P.~Wallis}, \bibinfo{author}{Z.~Allen-Zhu}, \bibinfo{author}{Y.~Li}, \bibinfo{author}{S.~Wang}, \bibinfo{author}{L.~Wang}, \bibinfo{author}{W.~Chen},
\newblock \bibinfo{title}{Lo{RA}: Low-rank adaptation of large language models}  (\bibinfo{year}{2022}). \URLprefix \url{https://openreview.net/forum?id=nZeVKeeFYf9}.
\bibitem[{Grammarly(2024)}]{grammarly}
\bibinfo{author}{Grammarly},
\newblock \bibinfo{title}{Grammarly handbook}  (\bibinfo{year}{2024}). \URLprefix \url{https://www.grammarly.com/handbook/}.

\end{thebibliography}

\newpage
\appendix
\section{Prompt Templates}
\label{appendix:prompt_template}

\begin{table}[ht]
    \caption{Prompt Template for List Questions}
    \centering
    \fbox{\begin{minipage}{0.98\linewidth}
        \textbf{Context}: The FGFR3 P250R mutation was the single largest contributor (24\%) to the genetic group;
        Syndromic craniosynostosis due to complex chromosome 5 rearrangement and MSX2 gene triplication\\
        \textbf{Question}: Which human genes are more commonly related to craniosynostosis?\\
        \textbf{Ideal answer}: The genes that are most commonly linked to craniosynostoses are the members of the Fibroblast Growth Factor Receptor family FGFR3 and to a lesser extent FGFR1 and FGFR2. Some variants of the disease have been associated with the triplication of the MSX2 gene and mutations in NELL-1. NELL-1 is being regulated bu RUNX2, which has also been associated to cases of craniosynostosis. Other genes reported to have a role in the development of the disease are RECQL4, TWIST, SOX6 and GNAS.\\
        \textbf{Exact answer}: FGFR3;FGFR2;FGFR1;MSX2;NELL1;RUNX2;RECQL4;TWIST;SOX6;GNAS\\
        \#\#\#\\
        \textbf{Context}: The current article presents a concise review of network theory and its application to the characterization of AED use in children with refractory epilepsy;\\
        Recent results suggest that LCM has a dual mode of action underlying its anticonvulsant and analgesic activity.\\
        \textbf{Question}: What are the main indications of lacosamide?\\
        \textbf{Ideal answer}: Lacosamide is an anti-epileptic drug, licensed for refractory partial-onset seizures. In addition to this, it has demonstrated analgesic activity in various animal models. Apart from this, LCM has demonstrated potent effects in animal models for a variety of CNS disorders like schizophrenia and stress induced anxiety.\\
        \textbf{Exact answer}: refractory epilepsy;analgesic;CNS disorders\\
        \#\#\#\\
        \textbf{Context}: \{context\} \\
        \textbf{Question}: \{body\} \\
        \textbf{Ideal answer}:
    \end{minipage}}
\end{table}

\begin{table}[ht]
    \caption{Prompt Template for Factoid Questions}
    \centering
    \fbox{\begin{minipage}{0.98\linewidth}
        \textbf{Context}: Ewing sarcoma is the second most common bone malignancy in children and young adults. It is driven by oncogenic fusion proteins (i.e. EWS/FLI1) acting as aberrant transcription factors that upregulate and downregulate target genes, leading to cellular transformation;\\
        Ewing sarcoma/primitive neuroectodermal tumors (EWS/PNET) are characterized by specific chromosomal translocations most often generating a chimeric EWS/FLI-1 gene\\
        \textbf{Question}: Which fusion protein is involved in the development of Ewing sarcoma?\\
        \textbf{Ideal answer}: Ewing sarcoma is the second most common bone malignancy in children and young adults. In almost 95\% of the cases, it is driven by oncogenic fusion protein EWS/FLI1, which acts as an aberrant transcription factor, that upregulates or downregulates target genes, leading to cellular transformation.\\
        \textbf{Exact answer}: EWS;FLI1 \\
        \#\#\#\\
        \textbf{Context}: Acrokeratosis paraneoplastica of Bazex is a rare but important paraneoplastic dermatosis, usually manifesting as psoriasiform rashes over the acral sites \\
        Bazex syndrome (acrokeratosis paraneoplastica): persistence of cutaneous lesions after successful treatment of an associated oropharyngeal neoplasm.\\
        \textbf{Question}: Name synonym of Acrokeratosis paraneoplastica. \\
        \textbf{Ideal answer}: Acrokeratosis paraneoplastic (Bazex syndrome) is a rare, but distinctive paraneoplastic dermatosis characterized by erythematosquamous lesions located at the acral sites and is most commonly associated with carcinomas of the upper aerodigestive tract.\\
        \textbf{Exact answer}: Bazex syndrome \\
        \#\#\#\\
        \textbf{Context}: \{context\} \\
        \textbf{Question}: \{body\} \\
        \textbf{Ideal answer}:
    \end{minipage}}
\end{table}

\begin{table}[ht]
    \caption{Prompt Template for Summary Question}
    \centering
    \fbox{\begin{minipage}{0.98\linewidth}
        \textbf{Context}: Hirschsprung disease (HSCR) is a multifactorial, non-mendelian disorder in which rare high-penetrance coding sequence mutations in the receptor tyrosine kinase RET contribute to risk in combination with mutations at other genes.\\
        \textbf{Question}: Is Hirschsprung disease a mendelian or a multifactorial disorder?\\
        \textbf{Answer}: Coding sequence mutations in RET, GDNF, EDNRB, EDN3, and SOX10 are involved in the development of Hirschsprung disease. The majority of these genes was shown to be related to Mendelian syndromic forms of Hirschsprung's disease, whereas the non-Mendelian inheritance of sporadic non-syndromic Hirschsprung disease proved to be complex; involvement of multiple loci was demonstrated in a multiplicative model.\\
        \#\#\#\\
        \textbf{Context}: Significant decreases and progressively higher reductions of systolic and diastolic blood pressure were observed after RSD.\\
        The RSD presents itself as an effective and safe approach to resistant hypertension.\\
        \textbf{Question}: What are the outcomes of Renal sympathetic denervation?\\
        \textbf{Answer}: Renal sympathetic denervation reduces heart rate, systolic and diastolic blood pressure.\\
        \#\#\#\\
        \textbf{Context}: \{context\} \\
        \textbf{Question}: \{body\} \\
        \textbf{Answer}
    \end{minipage}}
\end{table}

\clearpage

\section{Example LLM Prompt and Response}
\label{appendix:prompt_response}

\begin{table}[ht]
    \caption{Yes/No Question - Example Prompt and Response}
    \centering
    \begin{flushleft}
        \textbf{Prompt}
    \end{flushleft}
    \fbox{\begin{minipage}{0.98\linewidth}
        (examples)\\
        \textbf{Context}: No significant differences were observed, but FOLFOXIRI-bevacizumab achieved numerically superior survival outcomes versus FOLFOX-panitumumab. \\
        These observations suggest that although doublet chemotherapy plus anti-EGFRs remains the preferred treatment in patients with left-sided RAS/BRAF wild-type mCRC, FOLFOXIRI-bevacizumab is a valuable option able to provide similar, if not better, outcomes at the price of a moderate increase in toxicity and may be adopted based on patients' preference and potential impact on quality of life\\
        \textbf{Question}: Are FOLFIRINOV plus Bevacizuma the preferred therapy for left-sided RAS wild type advanced colorectal cancer with aggressive clinical presentation? \\
        \textbf{Ideal answer}:
    \end{minipage}}\\
    \begin{flushleft}
        \textbf{Response}
    \end{flushleft}
    \fbox{\begin{minipage}{0.98\linewidth}
        Yes, based on the observed survival outcomes, doublet chemotherapy plus anti-EGFRs remains the preferred treatment in patients with left-sided RAS/BRAF wild-type mCRC, but FOLFIRINOV plus Bevacizumab can be considered based on patients' preference and potential impact on quality of life.\\
        \textbf{Exact answer}: Yes\\
    \end{minipage}}
\end{table}

\begin{table}[ht]
    \caption{List Question - Prompt and Sample response}
    \centering
    \begin{flushleft}
        \textbf{Prompt}
    \end{flushleft}
    \fbox{\begin{minipage}{0.98\linewidth}
        (examples)\\
        \textbf{Context}: Recent studies have indicated associations between aberrant behavior of amyloid-beta and tau proteins and various neurological diseases, such as Alzheimer's disease, Parkinson's disease, and amyotrophic lateral sclerosis, as well as retinal neurodegenerative diseases like Glaucoma and age-related macular degeneration. Additionally, these proteins have been linked to cardiovascular disease, cancer, traumatic brain injury, and diabetes.\\
        \textbf{Question}: Amyloid-$\beta$ is associated with what diseases?\\
        \textbf{Ideal answer}:
    \end{minipage}}\\
    \begin{flushleft}
        \textbf{Response}
    \end{flushleft}
    \fbox{\begin{minipage}{0.98\linewidth}
        Amyloid-$\beta$ is associated with Alzheimer's disease, Parkinson's disease, amyotrophic lateral sclerosis, Glaucoma, age-related macular degeneration, cardiovascular disease, cancer, traumatic brain injury, and diabetes.\\
        \textbf{Exact answer}: Alzheimer's disease; Parkinson's disease; amyotrophic lateral sclerosis; Glaucoma; age-related macular degeneration; cardiovascular disease; cancer; traumatic brain injury; diabetes\\
    \end{minipage}}
\end{table}

\end{document}